\documentclass[10pt,twocolumn,letterpaper]{article}

\usepackage[pagenumbers]{cvpr} 

\definecolor{cvprblue}{rgb}{0.21,0.49,0.74}
\usepackage[pagebackref,breaklinks,colorlinks,allcolors=cvprblue]{hyperref}

\def\paperID{6} 
\def\confName{AI4CC}
\def\confYear{2025}

\usepackage{booktabs}
\usepackage{threeparttable}
\usepackage{multirow}
\usepackage{amsmath}
\usepackage[noend]{algpseudocode}
\usepackage{hyperref} 

\newcommand{\OURNAME}{EOPose\hskip0.25em}

\title{\OURNAME: Exemplar-based object reposing using Generalized Pose Correspondences}

\author{Sarthak Mehrotra\textsuperscript{1}
\and
Rishabh Jain\textsuperscript{2}
\and
Mayur Hemani\textsuperscript{2}
\and
Balaji Krishnamurthy\textsuperscript{2}
\and
Mausoom Sarkar\textsuperscript{2}\\
\textsuperscript{1}Indian Institute of Technology, Bombay \quad \textsuperscript{2}MDSR Lab, Adobe
}

\begin{document}
\maketitle
\begin{abstract}
   Reposing objects in images has a myriad of applications, especially for e-commerce where several variants of product images need to be produced quickly. In this work, we leverage the recent advances in unsupervised keypoint correspondence detection between different object images of the same class to propose an end-to-end framework for generic object reposing. Our method, \OURNAME, takes a target pose-guidance image as input and uses its keypoint correspondence with the source object image to warp and re-render the latter into the target pose using a novel three-step approach. Unlike generative approaches, our method also preserves the fine-grained details of the object such as its exact colors, textures, and brand marks. We also prepare a new dataset of paired objects based on the Objaverse dataset to train and test our network. \OURNAME produces high-quality reposing output as evidenced by different image quality metrics (PSNR, SSIM and FID). Besides a description of the method and the dataset, the paper also includes detailed ablation and user studies to indicate the efficacy of the proposed method. (Dataset will be made public)
\end{abstract}

\section{Introduction}
\label{sec:intro}
With the recent developments in deep neural network-based image generation methods, the interest in automatically editing images is steadily increasing. A related problem is that of producing different views of an object which finds applications in e-commerce because brands need to showcase their products in different settings for the various online marketing channels, and it is prohibitively expensive to shoot individual photos of each product or to edit them manually using image editing tools. While it is relatively easy to produce these multiple views through 3D rendering, most brands do not possess 3D models of their products. In this work, we propose reposing of generic objects using exemplar-based novel view generation.
\begin{figure}
\begin{center}
  \includegraphics[width=\linewidth]{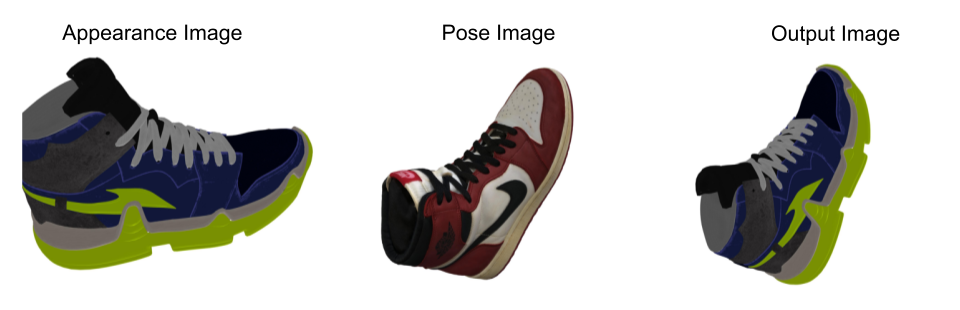}
\end{center}
\caption{Exemplar-based object reposing involves synthesizing an object in a desired pose denoted by another image of a similar kind of object}
\label{fig:tryon-problem}
\end{figure}

This problem has been explored extensively for certain classes like persons   ~\cite{li2019dense,liu2021neural,li2020pona,jain2023vgflow} and garments ~\cite{chopra2021zflow,viton,cpvton,jain2023umfuseunifiedmultiview}. These methods require a canonical definition of the pose of a human being either as a dense parametric representation ~\cite{guler2018densepose, loper2015smpl} or with sparse key points ~\cite{Cao_2017_CVPR, alphapose, yang2023effective}. However, they fail to generalize to other object classes because defining these pose representations for each class is infeasible. In \cite{zhou2018starmap}, Zhou et al. introduce a supervised method to identify class-agnostic object key points. However, the method produces a small set of key points inconsistent across different poses, and therefore not adequate for defining correspondence between the poses. We address this problem by specifying the desired target pose using an exemplar image of the same class as the object of interest and leveraging the sparse correspondences between the two images to produce the reposed output.

Another class of methods is based on denoising diffusion models (for example,  ~\cite{cao2024lightplane, cheng2023C3D}), attempts to solve the problem by learning a representation of the object from exemplar images and then conditionally generating the object pixels based on the pose. However, these methods are susceptible to hallucinating non-existent information and dropping vital information such as brand logos and textural patterns in their output limiting their usefulness for product images (Figure \ref{fig:qual1}). A related class of methods allows placing objects in arbitrary poses \cite{zhang2023controlcom, song2022objectstitch}, but they do not offer any control over the object's pose. Our method alleviates these problems and reduces hallucinations by employing a three-stage pipeline - the first stage identifies the key point correspondences between the source image and a target pose image, the second warps the source image based on the target pose, hence preserving details from the source image and the third stage conditionally re-renders the warped image to account for occlusions and to reconstruct missing details using an image-to-image GAN-like neural network. These steps preserve vital details in the source image and are less susceptible to hallucination except when the required information is not present in the source image.

Our contributions can be summarized as follows:
\begin{itemize}
    \item We formulate the problem of transferring the pose from one object image to another of a similar type in a class-agnostic manner (exemplar-based reposing)
    \item We propose \OURNAME, an end-to-end GAN-based pipeline that performs exemplar-based image reposing
    \item We provide extensive quantitative and qualitative results and a detailed user study, demonstrating significant improvements over various methods adapted to our problem statement
    \item We conduct ablation studies to analyze the impact of different design choices in \OURNAME\hskip-0.25em
\end{itemize}

\section{Related Work}
~\label{sec:related-work}
The problem of reposing objects is closely related to non-rigid object deformation such as for clothing (virtual try-on) and human reposing. Object compositing is another related problem where objects from images are introduced into new background settings in a certain pose. Since our method uses keypoints for matching and pose transformation, work done in pose representation is also a related problem. 

\subsection{Virtual Try-On}
There have been many developments in virtual try-on and human pose estimation. As compared to earlier methods (\cite{sekine2014virtual, pons2017clothcap}) that leveraged 3D scanners for virtual fitting of clothing items, recent methods \cite{jain2023vgflow,chopra2021zflow,cpvton,jain2023umfuseunifiedmultiview} directly use 2D images and synthesize a realistic image of a model from a reference image and an isolated garment image.  CP-VTON \cite{cpvton} uses a neural network to regress the transformation parameters of the TPS. Further, SieveNet \cite{sievenet} improves over \cite{cpvton} by estimating TPS parameters over multiple interconnected stages and also proposes a conditional layout constraint to better handle pose variation, bleeding, and occlusion during texture fusion. Another approach, ZFlow \cite{chopra2021zflow} proposes an end-to-end framework containing a combination of gated aggregation of hierarchical flow estimates termed Gated Appearance Flow. VG Flow \cite{jain2023vgflow} further develops on \cite{chopra2021zflow} by using a visibility-guided flow module to disentangle the flow into visible and invisible parts for style manipulation and simultaneous texture preservation. More recent works like \cite{kim2023stableviton, Zhu_2024_CVPR_mmvto, wang2024mv} use diffusion models which might be more susceptible to hallucination. \cite{kim2023stableviton} and \cite{wang2024mv} uses CLIP \cite{radford2021learningtransferablevisualmodels} image encoder to encode the garments for conditioning in diffusion models. \cite{Zhu_2024_CVPR_mmvto} uses diffusion transformers \cite{peebles2023scalablediffusionmodelstransformers} along with separate encoders for garments and humans for conditioning to generate a final image. 

\subsection{Mask-Guided Object Composition Methods} 
These methods integrate objects into masked portions of images by adjusting geometry and color. \cite{su2022general} utilizes thin-plate spline (TPS) based image warping and a generator to transfer object pose between images. They use a spatial structural block to preserve spatial details and a texture style block to retain appearance. However, per-pixel mask-based methods depend on precise masks, which don't account for differences in the inserted object's size and shape. Forcing the network to fit an object within a per-pixel mask \cite{su2022general} can lead to significant shape artifacts, making it unsuitable for exemplar-based reposing. More recent approaches like \cite{song2022objectstitch, song2024imprint} present self-supervised frameworks with conditional diffusion models. These frameworks feature content adaptors for semantic extraction and diffusion modules for seamless object-background blending. \cite{zhang2023controlcom} combines image blending, harmonization, view synthesis, and generative composition in a single diffusion model with a two-stage fusion strategy for enhanced realism. However, both methods can generate images with distorted poses due to inaccuracies in spatial pose correspondences and tend to hallucinate other textural details. Thus, such methods compromise both structural and surface detail integrity. Our method relies on generalized key point-based correspondences, which serve as a better guidance for the network.

\subsection{3D Pose Representation} 
Recent studies in single-image 3D reconstruction have examined various methods, including voxel, point cloud, octree, surface, and volumetric representations \cite{Varol_2018_ECCV, loper2015smpl, guler2018densepose, zheng2019deephuman, jackson20183d}. While human pose estimation is well-studied, pose estimation or keypoint detection for generic objects needs more development. Related concepts, such as SIFT \cite{lowe1999object}, focus on identifying interest points based on low-level image statistics. Other methods include heatmap representation for feature matching \cite{georgakis2018endtoend} and the multi-peak heatmap approach used by StarMap \cite{zhou2018starmap}, which provides key points with associated features and 3D locations. However, these methods often need more points to describe the object's pose effectively.

\begin{figure*}[t!]
\begin{center}
  \includegraphics[width=\linewidth]{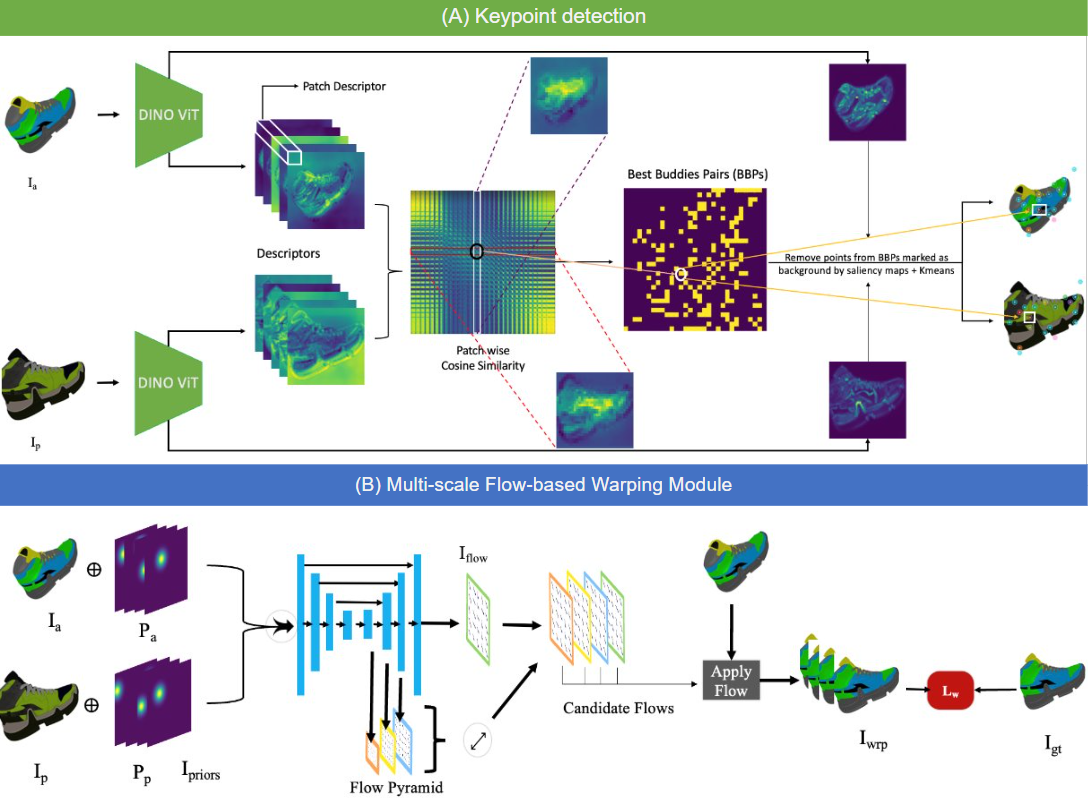}  
\end{center}
  \caption {A) Keypoint Detection: This module uses a pre-trained DINO-ViT to detect visual correspondences between the given pose image $I_p$ and appearance image $I_a$ B) Warping Module: This module takes in concatenated $I_a P_a, I_p, P_p$ as input and predicts a 2D flow used to deform the appearance image $I_a$ to align with the required pose denoted by pose image $I_p$}
\label{fig:warping_and_keypoint_diagram}
\end{figure*}

\section{Methodology}
\label{sec:method}
Given an image of the product in its initial pose called appearance image $I_a \in \mathbb{R}^{3\times H \times W}$ and another image showcasing the target pose called pose image $I_p \in \mathbb{R}^{3\times H \times W}$, \OURNAME transfers the pose from $I_p$ to $I_a$, generating the final output image $I_{gen} \in \mathbb{R}^{3\times H \times W}$. 

\begin{figure*}[h!]

\begin{center}

\includegraphics[width=0.9\linewidth]{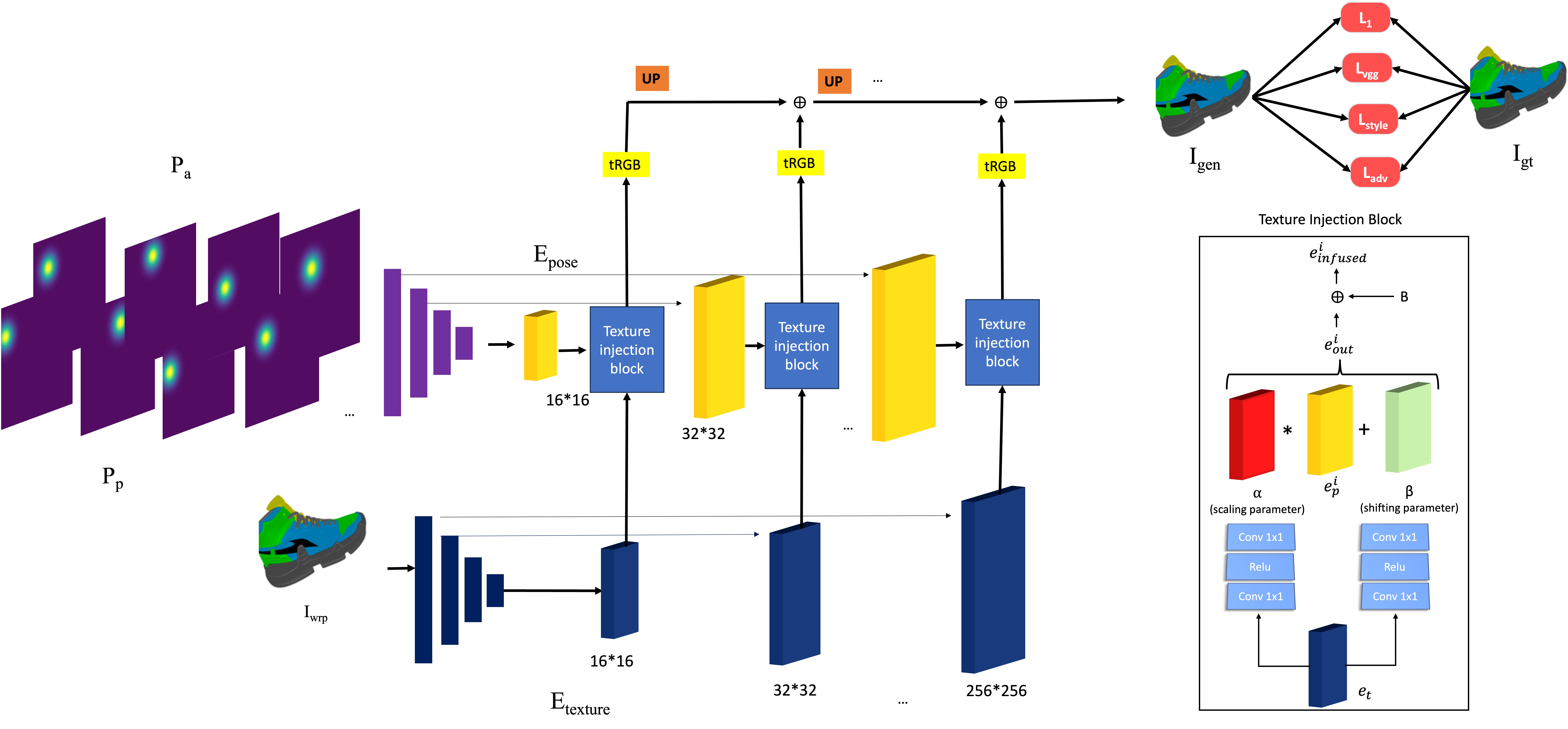}

\end{center}

\caption{Our Generator module consumes the warped image $I_{wrp}$ to generate the final reposed output. It utilizes a series of texture injection blocks to inject multi-scale appearance features into the pose encoding for the image generation process}

\label{fig:generator_diagram}

\end{figure*}

Figure \ref{fig:warping_and_keypoint_diagram} and \ref{fig:generator_diagram} depict the key ideas of the proposed solution. The problem of reposing an object present in a source image $I_a$ using the pose of an exemplar image $I_p$ is formulated as a three-step process:

\begin{enumerate}
\item \textbf{Keypoint Correspondence Detection:} Identifying Keypoint correspondences between the source image ($I_a$) which indicates the appearance of the object and the exemplar image ($I_p$) of the same class that supplies the target pose.

\item \textbf{Coarse Alignment:} Warping the source image $I_a$ to align with the pose exhibited in $I_p$, producing an intermediate warped image $I_{wrp}$.

\item \textbf{Fine-grained Re-rendering}: Generating the final reposed output $I_{gen}$ from the warped source image $I_{wrp}$ utilizing multi-scale appearance features and the pose encoding for the generation process.
\end{enumerate}
We discuss these next in some detail.


\subsection{Keypoint Correspondence Detection}
We first start with the keypoint detection process based on the method proposed by Amir et al. in \cite{amir2022deep}, which uses features obtained from a pre-trained DINO-ViT model. The method uses a pre-trained Vision Transformer (ViT) to extract Spatial feature descriptors $(S_a, S_p)  \in \mathbb{R}^{\frac{H}{8}*\frac{W}{8}\times 768}$ corresponding to the two input images $(I_a, I_p)  \in \mathbb{R}^{3\times H \times W}$, and salience maps from the attention module of the last layer of the ViT. 
It then computes cosine similarity between the descriptors of the two images and applies the ``Best Buddies Pairs'' (BBP)\cite{oron2016bestbuddiessimilarityrobust} approach to find the matching correspondences from both images. The matching is restricted to non-background patches by disregarding patches with zero salience map values. Finally, only mutually nearest neighbor descriptor pairs are retained. A pair of descriptors $s_a \in S_a$ and $s_p \in S_p$ are a best-buddy pair if and only if:
\begin{equation}
\text{NN}(s_a, S_p) = s_p \quad \text{and}\quad \text{NN}(s_p, S_a) = s_a
\end{equation}
where $\text{NN}(s_a, S_p)$ denotes the nearest neighbor of $s_a$ in set $S_p$ under cosine similarity.

Subsequently, K-means clustering is applied to the patch descriptors after concatenating them, where the desired number of points determines the number of clusters. Also, the patches are ranked based on the values in the salience maps, and the top-$k$ (where $k$ is predefined) points are selected for further processing. For our experiments, we set $k = 35$ points. These points are denoted as $P_a \in \mathbb{R}^{k\times2}$ (pose correspondences for $I_a$) and $P_p \in \mathbb{R}^{k\times2}$ (pose correspondences for $I_p$). The effect of this value on model performance is discussed in Section \ref{sec:ablations}.

\subsection{Coarse-Alignment with Pose-guided Warping}
The warping module (Figure \ref{fig:warping_and_keypoint_diagram}) is designed to transform the appearance image $I_a$ to align with the pose depicted in $I_p$ using a Skip-UNet \cite{UNet} architecture that estimates per-pixel warp parameters.
 We generate a 76-channel input for the U-Net by stacking the points along with their respective images. The input comprises of $I_a$ (3 channels), $P_a$ (35 channels), $I_p$ (3 channels), and $P_p$ (35 channels). The 2D points in \( P_a \in \mathbb{R}^{35 \times 2} \) and \( P_p \in \mathbb{R}^{35 \times 2} \) are converted into a 35-channel representation by encoding each point's \( x \) and \( y \) coordinates as Gaussian distributions centered at their respective locations, defined over the image dimensions \((H, W)\). This process results in a multi-channel input where each channel corresponds to one of the \( 35 \) points in an ordered fashion.
Now to get the image details, we pass it through our Skip-UNet network with 12 layers that processes the input of dimensions ($76,H,W$), generating $K$ candidate flow maps ($f_l$ for $l \in {0, ..., K-1}$), where each map $f_l$ is twice the size of its preceding map $f_{l-1}$ and $f_{K-1} \in \mathbb{R}^{2\times H \times W}$. These maps are then interpolated to a uniform size, creating a pyramid of $K$ maps with varying structural details. Following the approach in \cite{teed2020raftrecurrentallpairsfield}, all flow maps undergo convex upsampling, a generalization of bilinear upsampling that learns the upsampling kernel from the last U-Net layer and the given flow map. This technique enhances the map by preserving the smoothness and continuity of vectors, reducing artifacts, and retaining finer details. Finally, the warping module predicts per-pixel appearance flow, facilitating the transformation of the appearance image. This approach results in more accurate and realistic flow estimation, crucial for effectively warping $I_a$ to align with the pose in $I_p$.

\vspace{-3mm}
\paragraph{Image Warping} The output flow map $f_{K-1}$ is used to warp the appearance image $I_{a}$ to obtain the warped image $I_{wrp}$. Additionally, the intermediate flow maps $f_l$ for $ l \in \{0, .., K-2\}$ are also used to produce intermediate warped images ($I^{l}_{wrp}$).

\vspace{-3mm}
\paragraph{Losses} 

For each warped image, we apply three main losses: L1-loss ($L_{1}$), perceptual similarity loss ($L_{per}$), and style loss ($L_{sty}$). These losses are used at intermediate layers to regularize the flow module, enabling the network to learn both global and fine-grained warping details. The flow maps are also subjected to total variation loss ($L_{tv}(f_{l})$) and an initial TPS-based loss ($L_{flow}$) to ensure smooth transitions and quicker convergence (Details in supplementary).  
The combined warping loss, denoted as $L_{wrp}$ is the aggregate of these individual losses and is defined as:
\begin{small}
\begin{align}
\begin{split}
L_{wrp} &= \sum_{l=0}^{l=K-1}L_w(I^{l}_{wrp}, f_l)
\end{split}
\end{align}
\end{small}

for,
\begin{small}
\begin{align}
\begin{split}
L_w(I,f) &=\beta_{1} \|I, I_m^{gt}\|_1 + \beta_{2} L_{per}(I, I_m^{gt}) 
+ \beta_{3} L_{sty}(I, I_m^{gt}) \\ &\enskip + \beta_{4} L_{flow}(f, f_{tps}) + \beta_{5}L_{tv}(f)
\end{split}
\end{align}
\end{small}

\vspace{-5mm}

\subsection{Fine-grained Re-rendering}

The generator module (Figure \ref{fig:generator_diagram}) is a GAN \cite{gans} based module designed to improve the texture and design of the final warped image $I_{wrp}$. It takes as input the pose correspondences $P_p$, appearance correspondences $P_a$, and the final warped image $I_{wrp}$. The intermediate warped images are not used in the generator. 

\paragraph{Pose Encoder} The pose encoder is built upon a ResNet architecture and is designed to process both pose correspondences, denoted as \(P_p\), and appearance correspondences, denoted as \(P_a\). By leveraging both pose and appearance correspondences, the encoder facilitates a more robust understanding of the target image regions from which corresponding textures can be directly derived from the source image. This dual-input approach significantly mitigates the risk of generating unrealistic or hallucinatory textures. Furthermore, the network is trained to effectively identify and model the complex relationships between the two poses, enabling improved synthesis quality and alignment in downstream tasks.



\vspace{-2mm}
\paragraph{Texture Encoder} The texture encoder employs a ResNet architecture, similar to the pose encoder, to process \(I_{wrp}\), generating texture encodings across multiple hierarchical scales. Low-resolution features are designed to effectively capture the object’s overarching semantics and stylistic attributes, while high-resolution features preserve intricate, fine-grained details from the source image. To further enhance the representation, skip connections are incorporated into the texture encoder, enabling the seamless integration of low- and high-resolution features. This design ensures that multiple levels of semantics are captured, contributing to a more comprehensive and nuanced texture representation.



\paragraph{Texture Injection Block}The outputs from the texture encoder along with the pose encodings are passed through a texture injection block which integrates texture embeddings (\(e_t\)) into pose embeddings (\(e_p^i\)) using 2D style modulation \cite{albahar2021pose} to produce texture-infused embeddings (\(e_\text{infused}^i\)). This integration is achieved through a two-step process. First, the texture embedding (\(e_t\)) is processed using a series of two 1x1 convolutional layers, interspersed with ReLU activations (Figure \ref{fig:generator_diagram}), to generate scaling (\(\alpha\)) and shifting (\(\beta\)) parameters. These parameters are then applied to the pose embedding (\(e_p^i\)) through an element-wise scaling operation (\(e_p^i \times \alpha\)) followed by an additive shifting operation (\(+ \beta\))


\begin{equation}
    e^i_{\text{out}} = \alpha * e^i_p + \beta 
\end{equation}


The resulting output embedding (\(e_\text{out}^i\)) undergoes a noise modulation operation (\(B\)) to introduce controlled stochastic variations, ultimately producing the texture-infused embedding (\(e_\text{infused}^i\)). This operation helps the generator produce more realistic images.
\begin{equation}
    e^i_{out} = B(e^i_{\text{infused}}) 
\end{equation}

\vspace{-4mm}

\paragraph{tRGB Block} Each $e^i_{out}$ obtained from the texture injection block is processed by a tRGB block. The tRGB block consists of a $1\times1$ convolution layer with 3 output channels, producing the final image at a specific resolution. This output is then added using residual connections post bilinear upsampling to the size of the next block.

\vspace{-3mm}
\paragraph{Losses} We use L1, Perceptual, Style, and LSGAN losses between $I_{out}$ and $I_{gt}$. L1 loss preserves pixel-level identity and texture. Perceptual and Style losses ensure high-level semantic alignment. LSGAN loss, applied with the target pose $P_p$, improves pose alignment and enhances sharpness by reducing artifacts. LSGAN is preferred for its superior results and training stability over traditional GAN loss \cite{mao2016squares}. Overall, the loss function can be defined as:

\begin{small}
\begin{align}
\begin{split}
L_{gen} =\; & \alpha_{l1}||I_{out}, I_{gt}||_1 + \alpha_{per}L_{per}(I_{out}, I_{gt}) \\
           & + \alpha_{sty}L_{sty}(I_{out}, I_{gt}) + \alpha_{adv}L_{adv}(I_{out}, I_{gt}, P_p)
\end{split}
\end{align}
\end{small}

\subsection{Training}
\subsubsection{Dataset Preparation}

Due to the absence of a comprehensive training dataset demonstrating pose transfer between objects, we generated and utilized a novel dataset from the 3D models available in Objaverse \cite{objaverse}. We select multiple 3D models of objects and then manually filter out those comprising of a single mesh to prevent interference from surrounding objects. Objects with similar initial orientations are grouped. Finally, we are left with 8 models of shoes, 12 models of briefcases, 29 models of vases, and 11 models of file cabinets used for the creation of this dataset. All these 3D models exhibit distinct variations in texture, design, and color.

To create the training data, we select two models of the same class and orient them using two different sets of random Euler angles (ranging from $30^{\circ}$ to $180^{\circ}$) for the $x$, $y$, and $z$ axes. Each model is rotated using these two sets of angles, resulting in four different images. Let's denote the models as $M_1$ and $M_2$, and the two angle configurations as $c_1$ and $c_2$. This process produces the following four images: $M_{1c1}$,$M_{1c2}$,$M_{2c1}$ and $M_{2c2}$ (as illustrated in Figure \ref{fig:dataset_diagram}, More details can be found in supplementary).

\begin{figure}[h!]
\begin{center}
  \includegraphics[width=\linewidth]{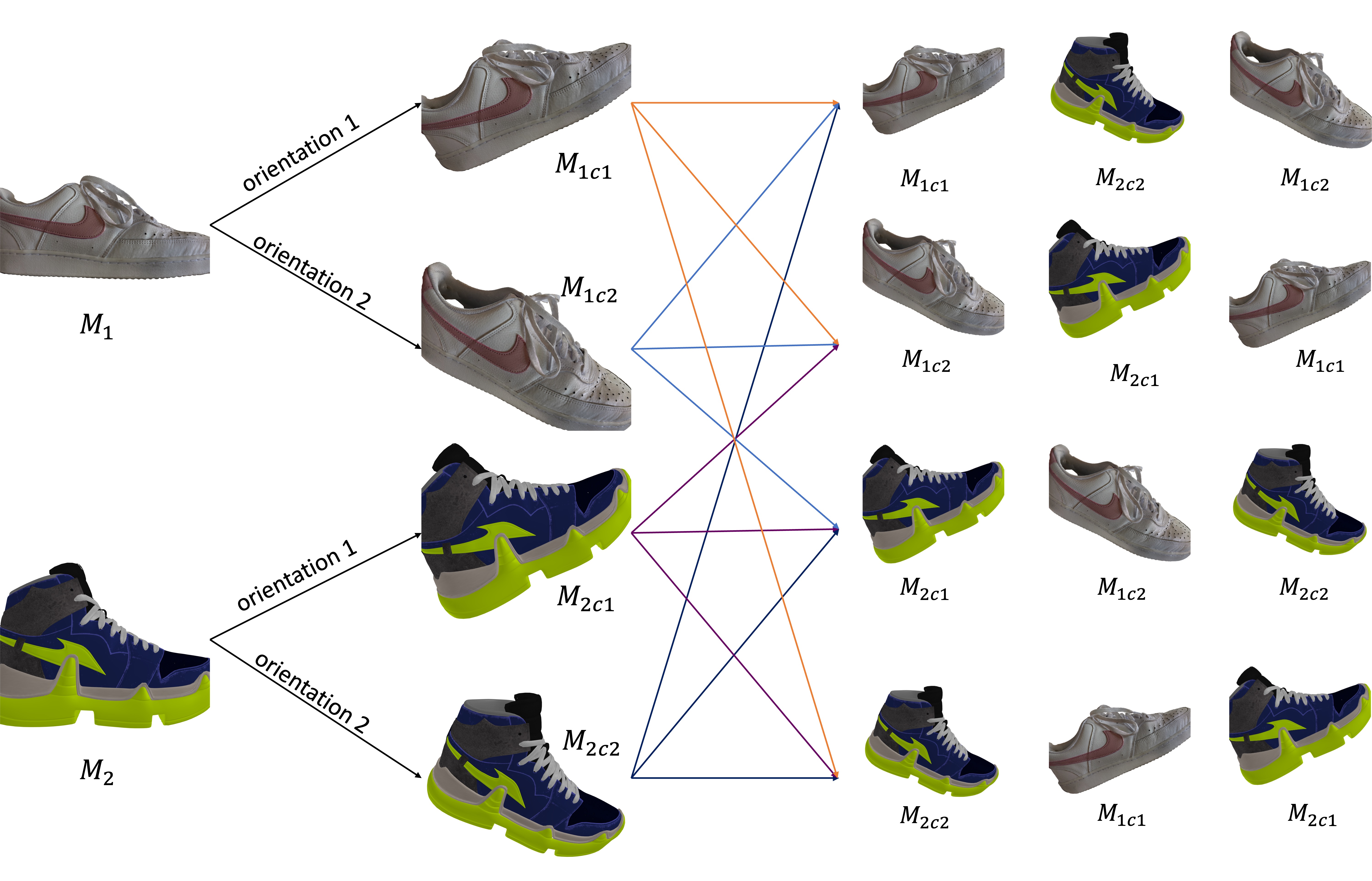}
\end{center}
  \caption{Paired dataset preparation by rendering Objaverse\cite{objaverse} 3d models of different objects in different poses for \OURNAME network training.}
\label{fig:dataset_diagram}
\end{figure}

\subsubsection{Training Hyperparameters}
We begin by training the warping module for a specified number of epochs. Post this training, we proceed to train the generator module for another set number of epochs, ensuring that the warping module remains frozen during this phase.
Following this, we optimize \OURNAME end-to-end with the following loss function:
\begin{equation}
    L_{total} = \alpha_{1}*L_{wrp}  + \alpha_{2}*L_{gen}
\end{equation}
where $\alpha_1, \alpha_2$  are scalar hyperparameters.
Other implementation details can be found in the supplementary.
\begin{table*}[h!]
\centering
\begin{threeparttable}
\resizebox{\textwidth}{!}{%
\begin{tabular}{lcccccccccccc}
\toprule
\textbf{Experiments} & \multicolumn{3}{c}{\textbf{TPS}} & \multicolumn{3}{c}{\textbf{UFO-PT}} & \multicolumn{3}{c}{\textbf{ControlCom}} & \multicolumn{3}{c}{\textbf{OURS}} \\
\cmidrule(lr){2-4} \cmidrule(lr){5-7} \cmidrule(lr){8-10} \cmidrule(lr){11-13}
 & SSIM$\uparrow$ & LPIPS$\downarrow$ & FID$\downarrow$ & SSIM$\uparrow$ & LPIPS$\downarrow$ & FID$\downarrow$ & SSIM$\uparrow$ & LPIPS$\downarrow$ & FID$\downarrow$ & SSIM$\uparrow$ & LPIPS$\downarrow$ & FID$\downarrow$ \\
\midrule
Vases & 0.68 & 0.28 & 42.81 & 0.27 & 0.36 & 200.2 & 0.23 & 0.67 & 300.36 & \textbf{0.89} & \textbf{0.04} & \textbf{11.82} \\
Briefcases & 0.61 & 0.35 & 213.52 &0.33&0.45&170.36& 0.24 & 0.66 & 160.56 & \textbf{0.86} & \textbf{0.06} & \textbf{39.68} \\
File Cabinets & 0.60 & 0.37 & 190.93 &0.21&0.68&237.82& 0.22 & 0.73 & 242.33 & \textbf{0.82} & \textbf{0.07} & \textbf{52.86} \\
Shoes & 0.55 & 0.27 & 58.84 & 0.35&0.52&160.84& 0.25 & 0.64 & 155.44 & \textbf{0.77} & \textbf{0.08} & \textbf{24.89} \\
All together & 0.34 & 0.58 & 75.22 & 0.30 & 0.48 & 133.9 & 0.23 & 0.69 & 147.20 & \textbf{0.44} & \textbf{0.33} & \textbf{18.22} \\
\bottomrule
\end{tabular}}
\caption{\OURNAME achieves significant improvement over existing baselines across different categories}
\label{tab:quant-sota}
\end{threeparttable}
\end{table*}

\section{Experiments}
In this section, we formalize the setup for our experiments for pose transfer. 
\vspace{-3mm}
\paragraph{Implementation Details} 
All experiments are conducted using PyTorch on Nvidia RTX 3090 GPUs. The warping module is trained with a batch size of 32 and the generator is trained using a batch size of 8. Both the modules are trained with a learning rate of 1e-4 using the Adam optimizer \cite{kingma2014adam}. 
\vspace{-3mm}
\paragraph{Evaluation Metrics} For \OURNAME, we use SSIM \cite{seshadrinathan2008unifying}, FID \cite{heusel2017gans}, and LPIPS \cite{zhang2018unreasonable} to evaluate generated outputs. We exclude the inception score (IS) based on the considerations outlined in \cite{barratt2018note}. SSIM measures image degradation by assessing luminance, contrast, and structure, making it valuable for our task where maintaining the object's structure is crucial. LPIPS evaluates patchwise similarity using deep learning model features and has shown a strong correlation with human perception of image similarity. This metric ensures that the features match on a patchwise level. FID  calculates the 2-Wasserstein distance between the InceptionNet statistics of the generated and ground truth datasets, serving as a reliable metric for assessing the realism of generated results. This serves to align the generated images with the original distribution to avoid out-of-distribution artifacts.
\vspace{-3mm}
\paragraph{Baselines} For \OURNAME, we compare performance using the outputs from Thin Plate Spline warping, UFO-PT \cite{su2022general} and ControlCom \cite{zhang2023controlcom}. We adapt our TPS method from \cite{cpvton}. For UFO-PT \cite{su2022general} and ControlCom \cite{zhang2023controlcom} we finetune the model using our dataset, perform inference using author-provided implementations, and present qualitative and quantitative comparisons.

\section{Results}
~\label{sec:results}
\begin{figure}[h!]
\begin{center}
  \includegraphics[width=\linewidth]{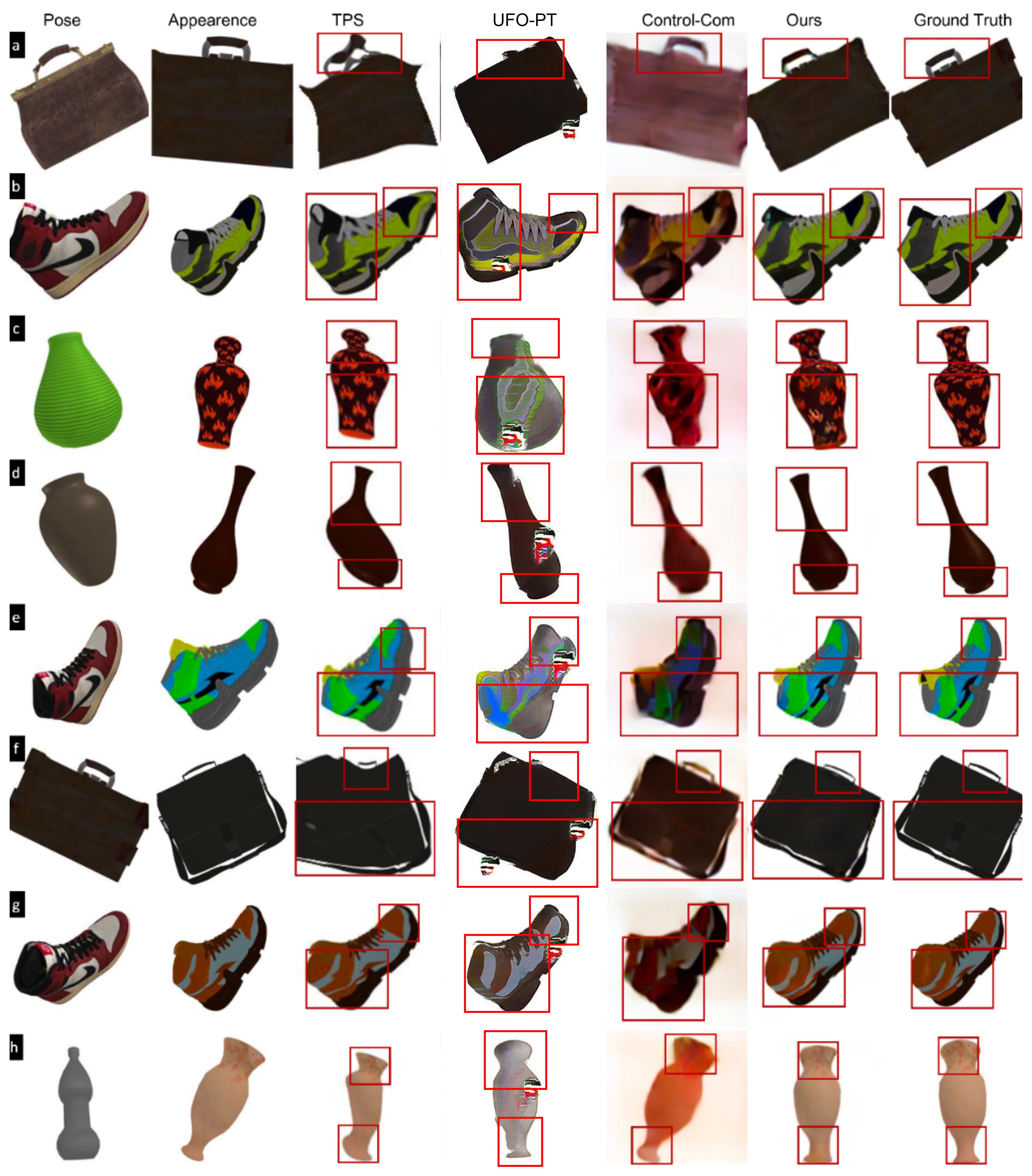}
\end{center}
\caption{In this figure we show improvements along different qualitative aspects compared to thin plate spline and ControlCom \cite{zhang2023controlcom}. We emphasize the differences in preserving pose (a,g,h), maintaining geometric integrity (b,e), and texture integrity (c,f,g,h).} 
\label{fig:qual1}
\end{figure}

We present quantitative (in Table~\ref{tab:quant-sota}) and qualitative results (Figure~\ref{fig:qual1}) along with a user study that highlights the superiority of \OURNAME over other baselines.

\vspace{-4mm}
\paragraph{Quantitative Results}
Table~\ref{tab:quant-sota} compares the performance of \OURNAME against state-of-the-art baselines for reposing generic objects. We report performance for Thin plate spline warping (TPS), UFO-PT \cite{su2022general} and diffusion-based approaches \cite{zhang2023controlcom}. In comparison to TPS and ControlCom\cite{zhang2023controlcom}, \OURNAME achieves significantly better SSIM of 0.44, LPIPS of 0.33 and FID of 18.22, compared to the next best values (SSIM=0.34, LPIPS=0.58 and FID=75.22). 


While thin plate spline (TPS) warping is constrained by the lack of degrees of freedom, leading to suboptimal warping quality, our method leverages dense flow prediction, allowing it to accurately preserve visible regions in the new pose while simultaneously performing style transfer for occluded or invisible regions. \cite{su2022general} uses mask guidance for both reference and target poses and employs a warping technique based on these masks. A diffusion-based approach is employed in \cite{zhang2023controlcom}, alongside mask guidance for reference poses similar to \cite{su2022general}. However, binary masks prove inadequate for defining poses, especially when rotations along all three axes are involved. Furthermore, the diffusion-based method used by \cite{zhang2023controlcom} can compromise objects' geometric and texture integrity in the generated images, with the denoising process introducing unwanted artifacts and leading to hallucinations. In contrast, our flow-based approach, \OURNAME, predicts the flow to warp the appearance image and then refines it with a generator, thereby reducing hallucinations.

\begin{figure}[h!]
    \centering
    \includegraphics[width=0.4\linewidth]{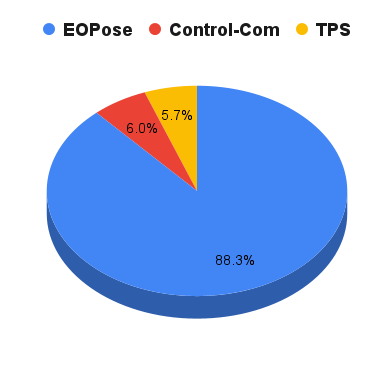}
    \caption{Survey results indicating the percentage of images where \OURNAME was preferred over competing methods.}
    \label{fig:user-study}
\end{figure}

\vspace{-4mm}
\paragraph{Qualitative Results}
Figure~\ref{fig:qual1} illustrates qualitative comparison with TPS, \cite{su2022general} and \cite{zhang2023controlcom}, the baselines with available code implementations. We contrast the final outputs along varying dimensions of quality. These include factors that determine the realism of the generated image as a whole and the local geometry, colors, and patterns.

In Figure \ref{fig:qual1} (a), \OURNAME generates images with accurate pose and texture, capturing intrinsic details like the part near the handle highlighted by the bounding box. In contrast, other works failed to maintain coherence during generation. TPS, \cite{su2022general}, and \cite{zhang2023controlcom} failed to reconstruct the highlighted parts correctly and align the pose of the appearance image with the pose image.

In (b), \OURNAME maintains the pattern and shape of the highlighted parts, showcasing the design on the shoe, which other works could not achieve. TPS and \cite{su2022general} struggle to maintain the structure of the highlighted regions correctly. In (c), \OURNAME preserves the pattern and texture fidelity of the vase's design, whereas \cite{zhang2023controlcom} fails to maintain the design consistency and \cite{su2022general} is not able to generate the results.

In (d), \OURNAME accurately preserves the texture and colors of the output, unlike \cite{zhang2023controlcom}, which introduces a lighter shade. TPS and \cite{su2022general} also fail to attain the correct pose and texture and introduce significant artifacts. In (e), \OURNAME captures sharp color transitions, consistent shape, and small design details, closely aligning with the expected output. Previous methods exhibit deformation and blurriness. Although TPS is able to attain the correct pose, it compromises the shoe's structure. \cite{su2022general} also introduces distortions and fails to maintain fine details.

In (f), \cite{zhang2023controlcom} slightly alters the design and color of the bag handle. This artifact is missing from \cite{su2022general}. In contrast, \OURNAME accurately replicates these details, including capturing the proper hole shape at the top, while maintaining the original color and reducing hallucinations. In (g), \cite{su2022general} and \cite{zhang2023controlcom} misinterpret the pose and fail to reproduce the pattern. However, \OURNAME successfully captures the pose and maintains the design, demonstrating robustness in handling complex poses.

Lastly, in (h), TPS, \cite{su2022general}, and \cite{zhang2023controlcom} distort the color and shape of the vase, whereas \OURNAME preserves the pattern at the top of the vase and maintains the shape to a greater extent.

\paragraph{User Study}
We conducted a survey with 30 volunteers from 7 institutions, encompassing diverse ages, genders, and occupations. Each volunteer reviewed 20 randomly selected results from a test set of 800 samples, comparing \OURNAME with TPS and ControlCom \cite{zhang2023controlcom}. For each comparison, volunteers saw the pose and appearance images along with the three generated results. Volunteers selected the best output without a time limit. The results, shown in Figure \ref{fig:user-study}, indicate a strong preference for \OURNAME, with 88.4\% of participants favoring our results over the baselines.
\begin{table}[h!]
\begin{center}
\begin{tabular}{|p{3cm}|c|c|c|}
\hline
\textbf{Experiments} & \textbf{SSIM} $\uparrow$ & \textbf{LPIPS} $\downarrow$ & \textbf{FID} $\downarrow$ \\
\hline
\centering15 keypoints & 0.42& 0.49 & 21.62 \\
\centering25 keypoints & 0.43 & 0.50 & 20.65 \\ 
\centering35 keypoints & 0.44& 0.33  & 18.22 \\ 
\centering45 keypoints & 0.45 & 0.54 & 22.48 \\
\hline
\centering Without $I_p$ & 0.39 &0.57 & 22.52 \\ \hline
\centering Without End-to-End &0.34&0.53 &34.86  \\ \hline
\centering\textbf{\OURNAME (OURS)} & \textbf{0.44} & \textbf{0.33} & \textbf{18.22} \\ \hline
\end{tabular}
\end{center}
\caption{Ablation studies for various design choices for object warping and generator in \OURNAME}
\label{tab:ablations-orpose}
\end{table}

\vspace{-15pt}
\section{Ablation Studies}
\label{sec:ablations}
In this section, we analyze the impact of different decisions and summarize results in Table~\ref{tab:ablations-orpose}.

\subsection{Number of Correspondence Points}
\label{sec:ablations-num_points}
\begin{figure}[h!]
\begin{center}
  \includegraphics[width=\linewidth]{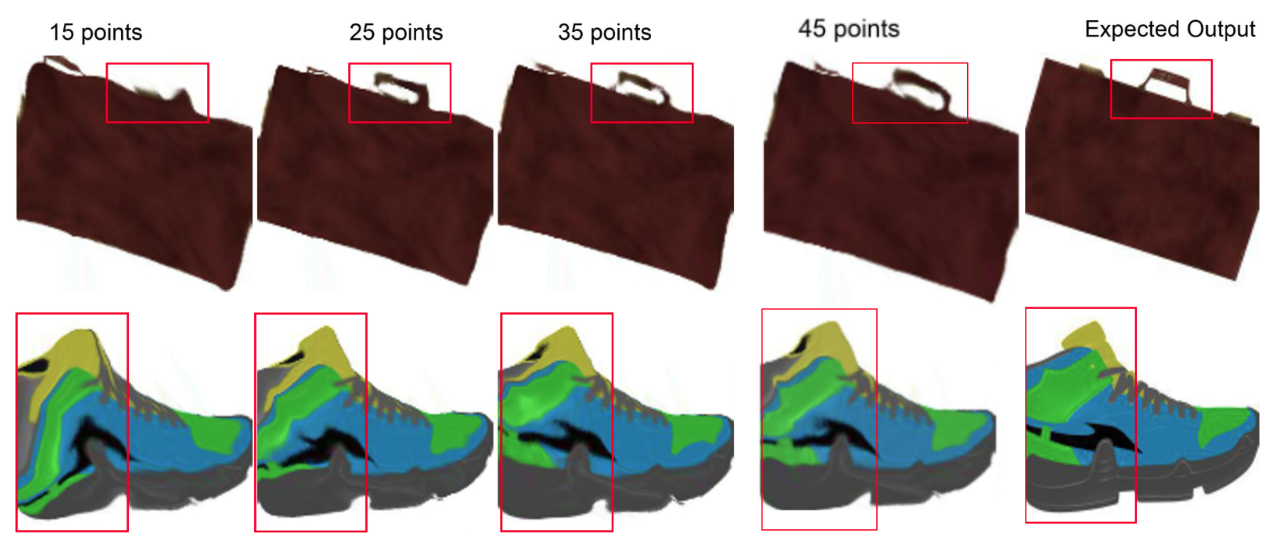}
\end{center}
\caption{Comparison of images after warping using flows from the warping module with different keypoint counts} 
\label{fig:figure_different_points}
\end{figure}

First, we illustrate how the number of correspondence points affects the generated flow and the resulting image. Reducing the number of correspondence points diminishes the flow quality, as the model has less information to generate the flow accurately and must infer or guess certain aspects. As shown in Table \ref{tab:ablations-orpose} the SSIM decreases from 0.44 to 0.42 and the FID increases from 18.22 to 21.62 when the number of keypoints is reduced from 35 to 15. Figure \ref{fig:figure_different_points} shows the flows obtained from different objects using different keypoint numbers. In Figure \ref{fig:figure_different_points}, row 2 demonstrates that increasing the number of points improves the reconstruction quality of specific components, such as the region near the heel of the shoe. Row 1 shows better artifact regeneration, like the handle on the briefcase. However, more keypoints introduce noise beyond a certain threshold and confuse the network, as observed in Table \ref{tab:ablations-orpose}, rows 1-4. We found that using 35 points yields satisfactory results.

\subsection{Input Priors and Training}
\label{sec:ablations-input-priors}
\begin{figure}[h!]
\begin{center}
  \includegraphics[width=\linewidth]{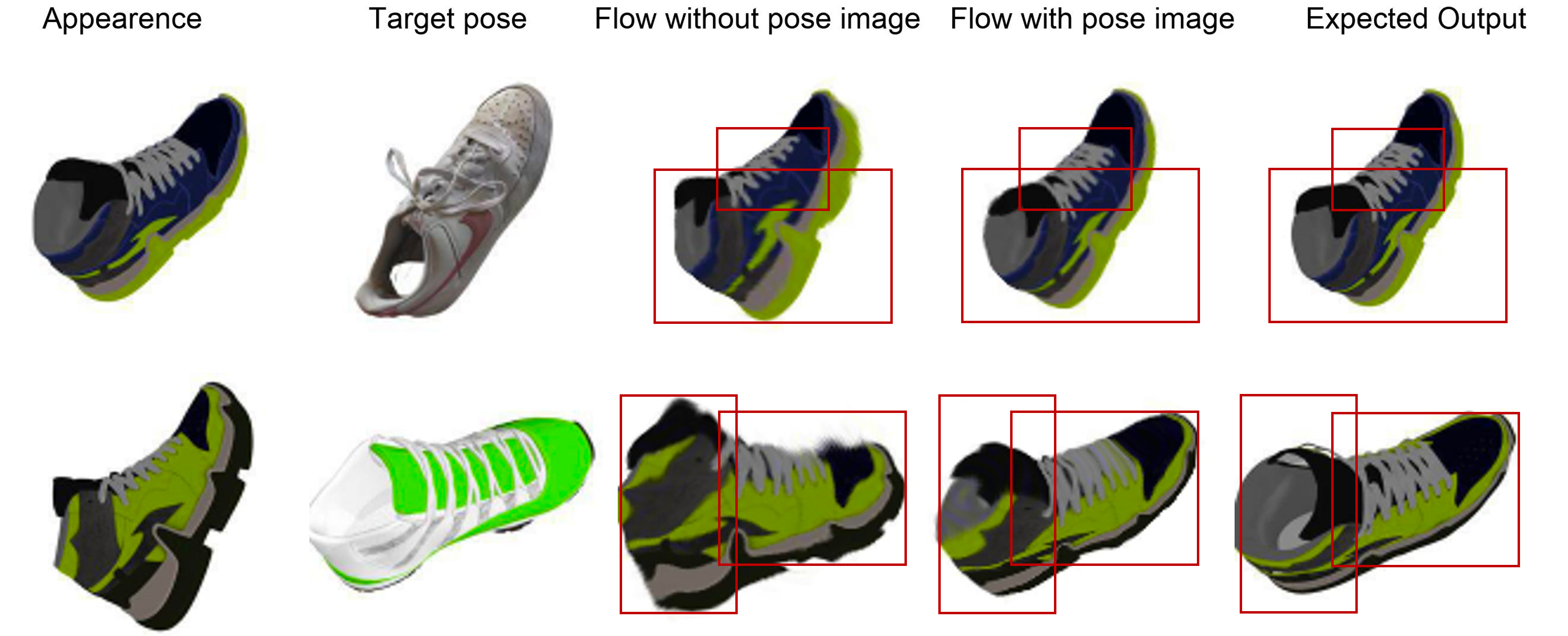}
\end{center}
\caption{This figure illustrates the importance of the pose image $I_p$ as an input to the warping module} 
\label{fig:figure_without_pose_image}
\end{figure}

\paragraph{Pose Image ($I_p$)} Along with the target pose keypoints $P_p$, the pose image $I_p$ is also provided to the warping module. This addition offers the module a clearer understanding of the desired target flow due to extra appearance information. As illustrated in Figure \ref{fig:figure_without_pose_image}, there is a noticeable difference in the output flows when the pose image is included in the warping module compared to when it is not.
Figure \ref{fig:figure_without_pose_image} demonstrates the image warped using the flows produced by warping modules trained with and without the pose image ($I_p$). Row 1 indicates that incorporating the pose image allows the model to have a clearer understanding of the structure and pose of the object. Row 2 illustrates that the pose image enhances the model's comprehension of the relationship between different components of $I_p$ and $I_a$, resulting in more accurate and coherent warping.

\vspace{-5mm}
\paragraph{End-to-end Fine Tuning} The end-to-end fine-tuning of the entire \OURNAME network (including the warping and generator modules) improves SSIM (0.34 to 0.437), LPIPS (0.53 to 0.33) and FID (from 34.86 to 18.22) of the try-on output as indicated in Table~\ref{tab:ablations-orpose} (row 6 vs 7).

\section{Conclusion}

We propose a novel problem statement of exemplar-based object reposing, where the goal is to transfer the pose from a given pose image to an appearance image. To address this, we introduce \OURNAME, an end-to-end reposing framework designed to transform an appearance image while maintaining the texture and geometric integrity. Additionally, we created a new dataset comprising 8800 object pairs in different poses specifically for training and testing this model. We demonstrate the effectiveness of \OURNAME by comparing state-of-the-art methods and extensive ablation studies.

{\small
\bibliographystyle{ieeenat_fullname}
\bibliography{egbib}
}

\end{document}